%% file: main.tex
\documentclass{article}
\usepackage[margin=1in]{geometry}

\input{math_commands.tex}

\RequirePackage{fancyhdr}
\usepackage[authoryear,round]{natbib}
\usepackage{hyperref}
\usepackage{url}
\usepackage{graphicx}
\usepackage{tikz}
\usepackage{cleveref}
\usepackage{booktabs}
\usepackage{tabularx}
\usetikzlibrary{positioning,shapes,arrows.meta,intersections}
\usepackage{wrapfig}
\usepackage{ragged2e}
\usepackage{amssymb}
\title{Does Feedback Alignment Work at Biological Timescales?}
\setcitestyle{semicolon}
\usepackage{parskip}

\author{
  Marc Gong Bacvanski$^{1,2}$ \quad
  Liu Ziyin$^{1,2}$ \quad
  Tomaso Poggio$^{1}$ \\
  \\
  $^{1}$ Massachusetts Institute of Technology \\
  $^{2}$ NTT Research \\
}

\usepackage{graphicx}
\usepackage{booktabs}
\usepackage{tabularx}
\usepackage{wrapfig}

\usepackage{amsmath,amssymb}
\usepackage{cleveref}   
\usepackage{tikz}
\usetikzlibrary{positioning,shapes,arrows.meta,intersections}

\newcommand{\tprop}{\tau_{\mathrm{prop}}}
\newcommand{\tplas}[1]{\tau_{\mathrm{plas}}^{#1}}
\newcommand{\tdec}[1]{\tau_{\mathrm{dec}}^{#1}}

\begin{document}
\date{} 

\maketitle

\begin{abstract}
\noindent 
Feedback alignment and related weight-transport--free algorithms are often proposed as biologically plausible alternatives to backpropagation, yet they are typically formulated in discrete phases with implicitly synchronized forward and error signals. 
We develop a continuous-time model of feedback-alignment--type learning in which neural activities and synaptic weights evolve together under coupled first-order dynamics with distinct propagation, plasticity, and decay time constants. 
We show that learning is governed by the temporal overlap between presynaptic drive and a locally projected error signal, providing an analytic explanation for robustness to moderate timing mismatch and for failure when mismatch eliminates overlap. 
Our results show that in order for feedback-alignment--type algorithms to work at biological timescales, they must obey the same temporal overlap principle that applies to other biological processes like eligibility traces.
\end{abstract}

\section{Introduction}
Understanding how biological circuits learn has long been a central challenge at the interface of neuroscience and machine learning. Among the many proposals for biologically plausible learning, a prominent class consists of feedback-alignment--style\footnote{In this work, we will use the term ``feedback alignment'' to refer to the class of feedback-alignment--style algorithms.}, weight-transport--free algorithms such as feedback alignment (FA; \cite{lillicrap2016random}), direct feedback alignment (DFA; \cite{nokland2016direct}), and Kolen–Pollack / weight mirrors (KP; \cite{kolen1994backpropagation, akrout2019deep}). These algorithms aim to relax the strong requirements of backpropagation, such as exact weight transport, while preserving the ability to optimize deep networks. These algorithms are attractive both as potential models of cortical learning and as approaches for training in neuromorphic or analog hardware.

There are two primary difficulties in connecting FA--type algorithms to biology: learning in biology unfolds continuously in time, and does not rely on separate learning and inference phases. 
Realistic neurons operate with finite conduction and integration times, so neither inference, error propagation, nor plasticity can be assumed to occur instantaneously or in synchronized steps~\citep{kandel2000principles}. Unlike digital learning algorithms, biological systems do not alternate between distinct inference and learning phases. By contrast, feedback alignment is cast in discrete steps that alternate between inference and learning phases, effectively assuming that forward and backward signals are globally and instantaneously synchronized~\citep{lillicrap2016random,nokland2016direct,akrout2019deep,scellier2017equilibrium}. 
This leaves a basic question unaddressed: \emph{Do FA/KP/DFA--style learning rules still work when implemented as continuous-time processes with biologically realistic propagation and plasticity timescales?}

In this work, we address this gap directly by constructing a continuous-time model for feedback-alignment--style algorithms. In this model, both neural states and synaptic weights evolve according to coupled first-order differential equations. Each neuron receives a feedforward drive and a modulatory error drive, and updates its synapses through a local two-signal rule. Crucially, inference and learning occur simultaneously: there is no global phase separation, as all state variables evolve under the same ODE. The model is governed by distinct time constants for fast signal propagation, intermediate potentiation, and slow synaptic decay, an ordering that mirrors the hierarchy of timescales in real neural tissue. Notably, feedback alignment algorithms are inherently non-spiking, and so our continuous-time model also does not attempt to model spikes.

Our model reveals three main insights about feedback alignment in continuous time:
\begin{enumerate}
    \item \textbf{Continuous-time realization.} Feedback alignment works in continuous time: inference and learning can proceed simultaneously under coupled differential equations without explicit forward/backward phases.
    \item \textbf{Biologically realistic regimes.} Effective learning occurs when propagation, plasticity, and decay constants satisfy the hierarchy of timescales $\tau_{\mathrm{prop}} \ll \tau_{\mathrm{plas}} \ll \tau_{\mathrm{dec}}$, and align with biologically observed ranges.
    \item \textbf{Overlap as the operative principle.} Learning is governed by temporal overlap between presynaptic drive and error signals; when delays eliminate overlap, updates become biased and performance collapses.
\end{enumerate}

\section{Related Work}
There has been significant interest in biologically plausible models of learning. Our focus here is on feedback-alignment–type algorithms, but many related frameworks have been explored. Continuous-time learning rules arise in contrastive Hebbian learning \citep{xie2003equivalence}, equilibrium propagation \citep{scellier2017equilibrium}, and broader energy-based formulations \citep{hopfield1982neural,bengio2015early}. Latent-equilibrium networks and related models derive neural and synaptic dynamics as gradient flows, enabling approximations to backpropagation (and backpropagation through time) through slow continuous dynamics \citep{haider2021latent,ellenberger2024backpropagation}. While these approaches take other perspectives of learning as dynamical evolution, we concentrate on error-propagation rules---feedback alignment (FA), direct feedback alignment (DFA), and Kolen–Pollack (KP)---and their continuous-time realizations.

Biologically motivated approximations to backpropagation have also been surveyed extensively \citep{whittington2019theories}, including predictive-coding and dendritic-error models \citep{whittington2017approximation,sacramento2018dendritic}. These frameworks typically express neuronal dynamics in continuous time and define learning via locally computed error signals. Our work complements this literature by focusing specifically on weight-transport--free error-propagation algorithms and analyzing their formulation as coupled neural and synaptic ODEs in layered feedforward networks.

Continuous-time neural computation has also been studied in neural ODEs \citep{chen2018neural}, where network computation is expressed as an ODE over hidden states and gradients are obtained via adjoint methods or solver differentiation \citep{baydin2018automatic}. In contrast, we consider a single forward-in-time dynamical system in which both neural states and parameters evolve continuously, with weights updated online through locally propagated error signals.

Recent work on heterosynaptic circuits has demonstrated that networks with local plasticity rules linking forward and feedback pathways can approximate backpropagation on standard benchmarks \citep{liao2024self}. These results further motivate feedback-style algorithms (FA, DFA, KP) as promising candidates for biologically grounded and hardware-compatible learning.

Finally, the biological plausibility of feedback-alignment–type algorithms has been explored in several contexts. For example, \citet{lillicrap2020backpropagation} argued how FA may be realizable in the cerebellum, and \citet{koplow2025emergence} showed that such rules align with observed Hebbian and anti-Hebbian plasticity motifs. However, their behavior in fully continuous-time settings and at biologically realistic timescales remains insufficiently understood. Addressing this question is a central goal of the present work.

\section{A Continuous-Time Model of Feedback Alignment}\label{sec:model-formulation}
\begin{figure}
    \centering
    \resizebox{\linewidth}{!}{
        \input{figures/topologies}
    }
    \caption{Neuron design with two archetypal topologies. \textbf{At left (zoom):} continuous-time heterosynaptic neuron model whose weights evolve according to plasticity rules. Neurons receive forward input $\textbf{x}$ and produce activated output $z$ with weights $\textbf{w}$. Error signals $\textbf{e}$ enter through modulatory weights $\textbf{v}$ and drive plasticity of $\textbf{w}$. 
    \textbf{In center:} layerwise error propagation topology, where errors are propagated to previous layers via backward weights. \textbf{At right:} direct error propagation topology, where error signals are broadcast to previous layers directly.}
    \label{fig:neuron-and-topologies}
\end{figure}
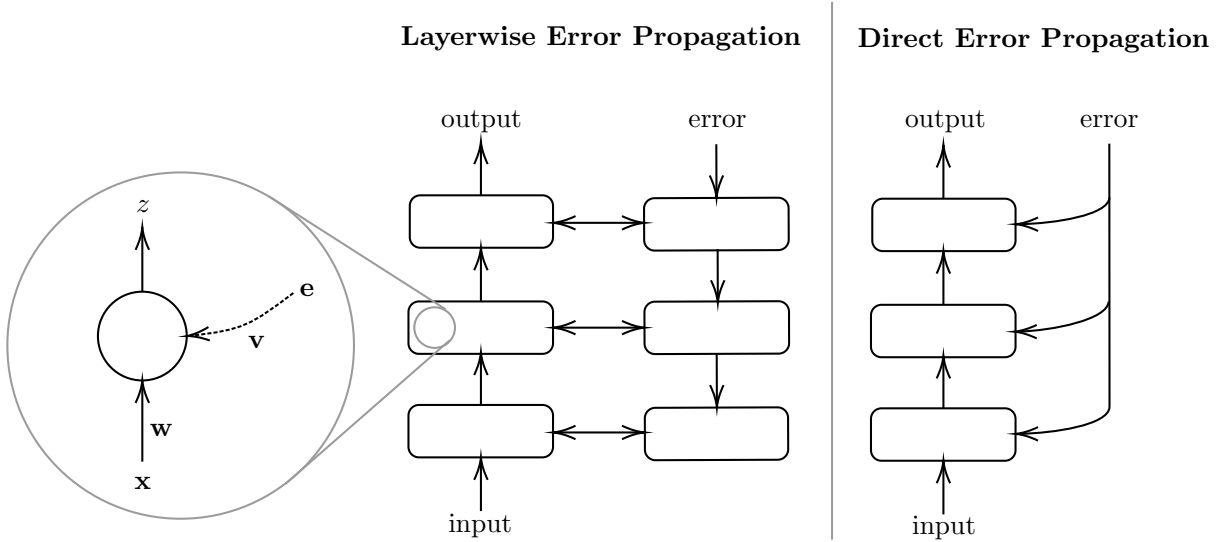


The FA--type algorithms feature a forward pathway that performs inference and a backward pathway that transmits error signals. While the original FA algorithm requires a fixed and randomly connected feedback pathway whose topology matches the feedforward, various variants have made the backward pathway much more flexible. For this reason, we study a broad and flexible set of connectivity topologies (see some discussions of this point in \cite{ziyin2025heterosynaptic}). Figure~\ref{fig:neuron-and-topologies} illustrates the continuous-time heterosynaptic neuronal systems we study. 
The forward pathway, parameterized by $\mathbf{w}$ and with activation $\sigma$, maps inputs $\mathbf{x}$ to neuronal output $z$. The error pathway, parameterized by $\mathbf{v}$, receives an error signal $\boldsymbol\epsilon$ and provides a modulatory influence on learning. Learning is heterosynaptic: the forward weights $\mathbf{w}$ update in proportion to the modulatory drive $(\mathbf{v}^\top \boldsymbol\epsilon)$, while the error weights $\mathbf{v}$ update in proportion to the forward drive $(\mathbf{w}^\top \mathbf{x})$. Let layer $l$ have width $d_l$. See Appendix~\ref{sec:appendix-neuron-dynamics} for the per-neuron update rules. Stacking the per-neuron variables into matrices, we can use the following notation: $\mathbf z_{l-1}(t)\in\mathbb R^{d_{l-1}}$, $\mathbf z_l(t)\in\mathbb R^{d_l}$, $W_l\in\mathbb R^{d_l\times d_{l-1}}$, and error source $\boldsymbol \epsilon_l(t)$ to define
\begin{align}
\dot{\mathbf z}_l = \frac{-\mathbf z_l + \sigma_l(W_l \mathbf z_{l-1})}{\tprop},
\quad
\dot W_l = -\frac{W_l}{\tdec{W}} + \frac{(V_l\boldsymbol\epsilon_l)\mathbf z_{l-1}^\top}{\tplas{W}},
\quad
\dot V_l = -\frac{V_l}{\tdec{V}} + \frac{(W_l \mathbf z_{l-1})\,\boldsymbol\epsilon_l^\top}{\tplas{V}}.
\label{eq:layer-dynamics}
\end{align}

Here $\tprop$ is the neuronal propagation time constant, setting how quickly $\mathbf z_l$ relaxes to its driven input.
The constants $\tplas{W},\tplas{V}$ control how rapidly synaptic plasticity occurs when presynaptic and modulatory drives coincide.
The constants $\tdec{W},\tdec{V}$ govern passive weight decay, setting the forgetting timescale. The center and right panels of Figure \ref{fig:neuron-and-topologies} show how these neurons connect into multilayer networks. Under timescale separation $\tprop{} \ll \tplas{} \ll \tdec{}$, the activation states equilibrate quickly, and weights update quasi-statically. Outputs track their steady state $\mathbf z_l\approx\sigma_l(W_l \mathbf z_{l-1})$ within a sample window. 

These dynamics can be understood as a continuous-time realization of the heterosynaptic two-signal principle. 
Under separation of timescales, we may treat neuronal activity as equilibrated relative to the slower synaptic updates. In this quasi-stationary regime, the fast variables satisfy $\mathbf z_l = \sigma(W_l \mathbf z_{l-1})$, and the weight dynamics reduce to the standard feedforward processing. At full stationarity ($\dot{\mathbf z}=\dot W=\dot V=0$) and with equal size, the fixed-point equations imply $W_l \propto (V_l\boldsymbol\epsilon_l)\mathbf z_{l-1}^\top$ and $V_l \propto (W_l \mathbf z_{l-1})\,\boldsymbol\epsilon_l^\top$, so that $W_l$ converges to a configuration consistent with $V_l^\top$. In this limit, the effective update to $W_l$ is the outer product of the input with the backpropagated error at that layer, recovering a backprop-like outer-product update. 
The effective weight update rule corresponding to different learning algorithms depends only on the definitions of the feedback weights $V_l$ and error drive $\boldsymbol{\epsilon}_l$. In Appendix \ref{appendix:discrete-time-limits} we discuss the discrete-time limits of these dynamics.

\begin{figure}
    \centering
    \begin{minipage}{0.48\linewidth}
        \centering
        \includegraphics[width=\linewidth]{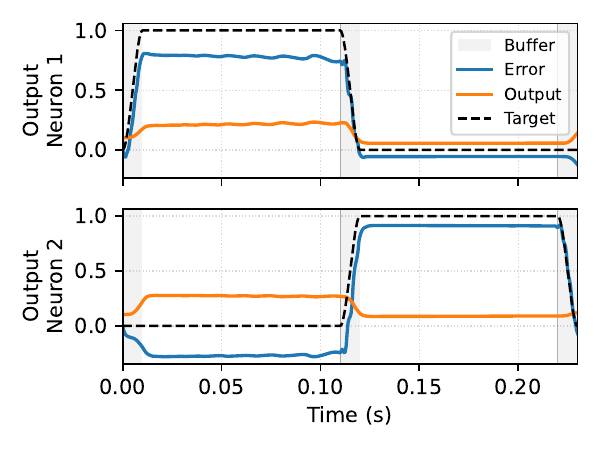}
        \caption{Zoomed-in view of output neuron dynamics during early training. The network input and corresponding error term changes during the \textit{buffer} time periods. Over a single sample, the neuron outputs move only minutely.}
        \label{fig:dynamics-early-training}
    \end{minipage}\hfill
    \begin{minipage}{0.48\linewidth}
        \centering
        \includegraphics[width=\linewidth]{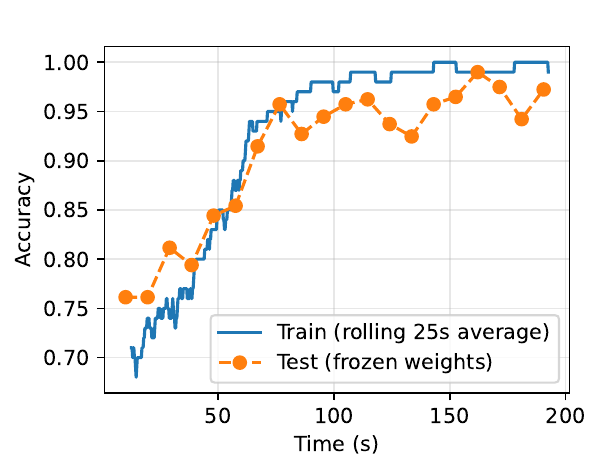}
        \caption{Sample train/test accuracy during training. Training set accuracy is measured as the moving average of instantaneous correct/incorrect predictions on the network input signals. Test set accuracy is measured by evaluating the network with frozen weight dynamics.}
        \label{fig:deep-mlp-train-vs-test}
    \end{minipage}
\end{figure}

We simulate training and inference of these models using ODE solvers~\citep{kidger2021on}. Inputs to the network are driven by the dataset's input signals, and the error at the output neurons in layer $L$ are $\mathbf e=\partial\mathcal L/\partial \mathbf z_L$. Figure~\ref{fig:dynamics-early-training} shows an example of the output-layer dynamics of early training. 
Each data point is presented to the network's input neurons for a fixed duration of time (sample time), with interpolation between the images done smoothly over a constant, much shorter duration (buffer time). Rather than impose the corresponding constraints and clamps on weights as traditional SGD/KP/FA/DFA prescribe, our experiments leave both $W$ and $V$ free to learn. We find that this makes our network architecture more general without sacrificing task performance.
During evaluation, parameters are held fixed and plasticity and decay dynamics are frozen: $\dot W=\dot V=0$ and $\boldsymbol\epsilon\equiv0$. Only inference dynamics $\dot {\mathbf z}$ are active. The model's prediction is read out at the end of the input presentation window, immediately before the next sample is shown.

\section{Continuous-time feedback alignment requires temporal overlap}\label{sec:timing-robustness}
\begin{figure}[h]
    \centering
    \includegraphics[width=\linewidth]{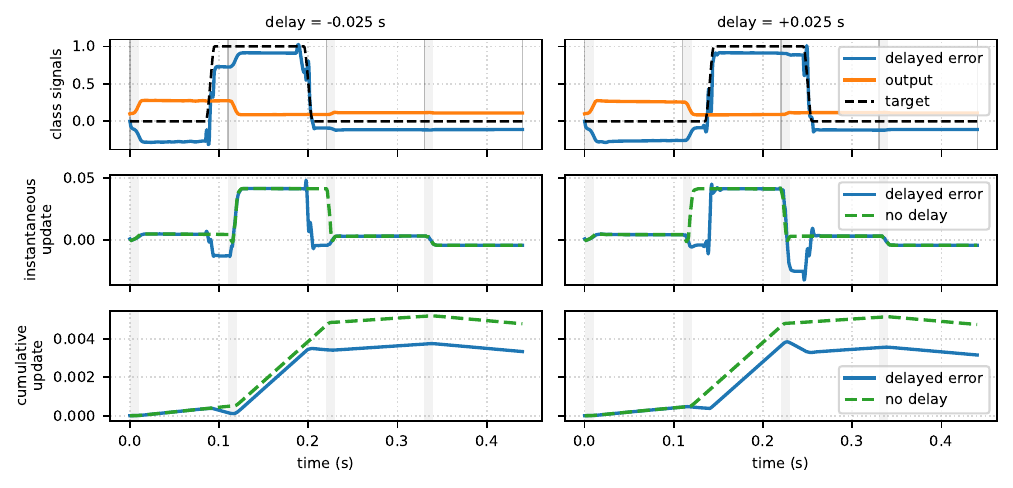}
    \caption{Single-neuron dynamics with different relative timings of the error and input signals. \textbf{Left:} error signal arrives early. \textbf{Right:} error signal is delayed. The bottom panel of both plots depicts the cumulative weight change and shows that in the presence of delay, the weight accumulates a biased gradient update compared to the case where there was no error delay.}
    \label{fig:error-signal-timing}
\end{figure}

A central question is: \emph{under what conditions does a synapse receive a correct update?} Two key considerations are (i) the temporal mismatch between input and error signals, and (ii) the rate at which the input changes. Figure~\ref{fig:error-signal-timing} illustrates the dynamics of a final-layer weight when the error signal arrives earlier or later than the corresponding input. During the mismatch period, the instantaneous weight update is incorrect, leading to a cumulative update that deviates from the non-delayed case.

To analyze robustness, we isolate only the part of the weight update that is informative: that which is proportional to the correlation between presynaptic activity and the matching error drive. Vectorized, the informative weight change accumulated over a presentation window of length $T$ is
\begin{equation}
\Delta W_l \;\propto\; \int_{0}^{T} \big(V_l \boldsymbol\epsilon_l(t)\big)\,\mathbf z_{l-1}(t)^\top\,k_{\tplas{}}(t)\;dt,
\qquad
k_{\tplas{}}(t)=\exp\!\big(-(T-t)/\tplas{}\big),
\label{eq:deltaW-kernel-vector}
\end{equation}
where the causal exponential kernel arises from the low-pass plasticity dynamics and weights more recent coincidence more strongly.
For a single synapse $(i \to j)$, this reduces to the scalar form
\begin{equation}
\Delta (W_l)_{j i} \;\propto\; \int_{0}^{T} (\mathbf z_{l-1}(t))_i \;\big((V_l)_{j, :}\, \boldsymbol\epsilon_l(t)\big)\;k_{\tplas{}}(t)\;dt.
\label{eq:deltaW-kernel-scalar}
\end{equation}
This makes clear that learning depends on the temporal cross-correlation between the presynaptic drive $(\mathbf z_{l-1})_i$ and the local modulatory/error drive at neuron $j$.

\textbf{Piecewise-constant inputs with delay.}
Assume $(\mathbf z_{l-1})_i(t)$ is active on $[0,T]$ and the error drive is active on $[\Delta,\Delta+T]$ (same duration, delayed by $\Delta$). In the fast-propagation limit $\tprop{}\ll\tplas{}$, the expected update becomes
\begin{align}
\mathbb{E}[\Delta (W_l)_{j i}]
&\propto
\int_{t_0}^{t_1} \exp\!\big((t-T)/\tplas{}\big)\,dt
\nonumber\\
&=\tplas{}\!\left(e^{-(T-t_1)/\tplas{}}-e^{-(T-t_0)/\tplas{}}\right)
=\tplas{}\,e^{-(T-t_1)/\tplas{}}\!\left(1-e^{-L/\tplas{}}\right),
\label{eq:exact-kernel}
\end{align}
where $t_0=\max(0,\Delta)$, $t_1=\min(T,\Delta+T)$, and $L=t_1-t_0=(T-|\Delta|)_+$.

\textbf{Flat-kernel limit and prediction.}
When $T\ll\tplas{}$ (the regime standard in our experiments), $k_{\tplas{}}$ is approximately constant over $[0,T]$, so~\eqref{eq:exact-kernel} reduces to the symmetric triangular law
\begin{equation}
\mathbb{E}[\Delta (W_l)_{j i}]\;\propto\; (T-|\Delta|)_+.
\label{eq:triangular}
\end{equation}
Thus learning succeeds if and only if input and error \emph{overlap in time}, and it degrades sharply as $|\Delta|\to T$. When $T$ approaches $\tplas{}$, the exact expression \eqref{eq:exact-kernel} predicts a mildly \emph{skewed triangle} that up-weights late-arriving errors (positive $\Delta$) relative to equally early ones; the skew vanishes continuously as $\tplas{}/T\to\infty$.

\textbf{A fixed overlap budget.}
For delay $\Delta$, define the correct-overlap set $C(\Delta)=[0,T]\cap[\Delta,\Delta+T]$ of length $L=(T-|\Delta|)_+$ and mismatched set $I(\Delta)=[0,T]\setminus C(\Delta)$. With plasticity kernel $k(t)$, let
\[
K_C(\Delta)=\!\!\int_{C(\Delta)}\! k(t)\,dt,\quad
K_I(\Delta)=\!\!\int_{I(\Delta)}\! k(t)\,dt,
\]
so that $K_C(\Delta)+K_I(\Delta)=K_T:=\int_0^T k(t)\,dt$. Thus overlap and mismatch trade off under a fixed budget. In the flat-kernel case $k\equiv1$, this reduces to $L+(T-L)=T$, yielding $\mathbb{E}[\Delta(W_l)_{j i}]\propto L=(T-|\Delta|)_+$, i.e. accuracy improves with either larger $T$ or smaller $|\Delta|$.

Our analysis predicts and experiments confirm near-symmetry between early and late error in the flat-kernel regime. By contrast, biological plasticity is causally gated: synaptic activity first writes a short-lived eligibility trace, and only subsequently arriving modulatory/error signals consolidate it into weight change~\citep{yagishita2014critical}.
Thus, early error fails to drive learning even though late error can---an asymmetry our simplified model compresses into a symmetric overlap law. Incorporating an explicit eligibility gate would recover this asymmetry without altering our conclusions about temporal overlap and timescale separation; we leave such extensions for future work.

\section{Feedback alignment works at biological timescales}\label{sec:experiments}
Running these experiments is computationally demanding. Each training and evaluation run requires explicit forward integration of a high-dimensional, stiff system of coupled ODEs. Neuronal activations evolve on millisecond timescales, while meaningful learning unfolds over hundreds to thousands of seconds. This range of time scales spans six orders of magnitude, forcing solvers to take very small steps across very long horizons.

We implement our simulations in JAX~\citep{jax2018github} and use the \texttt{Diffrax} library~\citep{kidger2021on} for ODE integration. Although we do not rely on JAX’s autodifferentiation, its just-in-time compilation and efficient array operations provide substantial performance gains for these large-scale simulations. For integration we use Tsit5~\citep{tsitouras2011runge}, an explicit 5th-order Runge–Kutta method with an embedded 4th-order error estimate. This combination is accurate and stable without resorting to computationally expensive implicit methods. The embedded estimate enables adaptive step sizing under PID control, which we find essential for efficiently handling stiffness and long simulation horizons.

Some of the parameter sweeps reported here, particularly the heatmaps over error delay and sample duration, require over 500 compute hours to complete. This reflects the intrinsic difficulty of simulating continuous-time learning rules, where propagation, plasticity, and decay processes must be faithfully resolved simultaneously. These results represent a substantial computational effort and demonstrate that continuous-time neural networks can be simulated effectively despite these challenges.

In this section, we evaluate three experimental regimes. Section~\ref{sec:direct-error-routing} demonstrates that networks can robustly learn so long as error signals overlap with inputs. Section~\ref{sec:layerwise-error-routing} demonstrate that deeper networks accumulate propagation lag, making them less tolerant to error delays and requiring longer sample times for stable learning. Finally, Section~\ref{sec:biological-timescales} shows that our framework operates effectively under synaptic, plasticity, and decay timescales that align with known cortical physiology.

\subsection{Direct Error Routing}\label{sec:direct-error-routing}
\begin{figure}[h!]
    \centering
    \includegraphics[width=\linewidth]{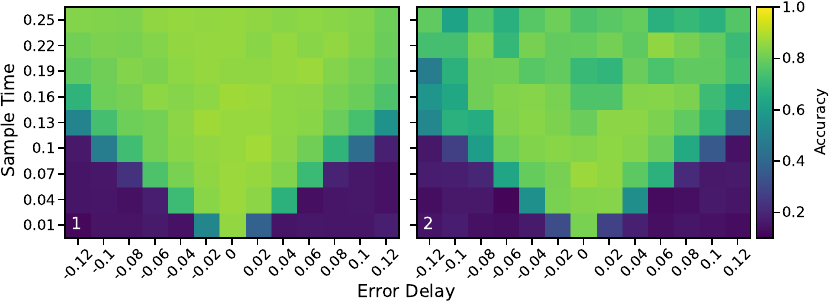}
    \caption{
    Evaluation of the direct error routing topology on the $7\times7$ downsampled MNIST dataset. The $x$-axis denotes the temporal delay between input signal and label: negative values indicate the label arrives before the input, while positive values indicate the label arrives after the input. At left is a network with 1 hidden layer of 49 neurons, and at right is a network with 2 hidden layers of 49 and 32 neurons. The learning is fairly robust until the delay exceeds the sample time.
    }
    \label{fig:dfa_heatmaps_delay_integration}
\end{figure}

The direct error routing topology sets the local error at each layer to be the global error (from the output layer), $\boldsymbol{\epsilon}_l=\mathbf e_L$. This corresponds to a Direct Feedback Alignment scheme where the backward weights $V_l$ are also learned. Figure~\ref{fig:dfa_heatmaps_delay_integration} examines the accuracy of the direct error routing topology on the $7\times7$ downsampled MNIST dataset as the error-signal delay and sample duration are varied. 

In the direct error routing topology, each layer receives a direct copy of the error signal. As a result, all neurons in the network receive the error signal with a propagation delay of one synaptic length. Learning fails when the error signal delay is approximately equal to or longer than the sample duration, since the error signal has zero overlap with its corresponding input. In these regions (accuracy $\sim10\%$), weights only change due to spurious correlations between mismatched input-label pairs. Transitional regions ($\sim70\%$ accuracy) arise when error and input partially overlap: some useful updates occur, but they are mixed with incorrect updates from mismatched periods. These networks have a fixed $\tplas{}=10$s while the sample time $T$ lies in 0.01-0.25s, putting them primarily in the flat-kernel learning limit. As a result, learning responds fairly symmetrically to both early and late error signals.

\subsection{Layerwise Error Routing}\label{sec:layerwise-error-routing}
\begin{figure}
    \centering
    \includegraphics[width=1\linewidth]{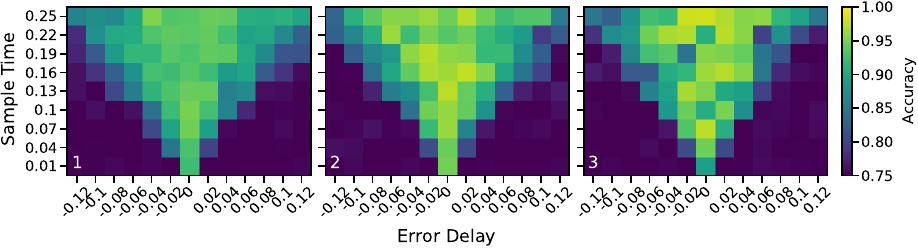}
    \caption{Evaluation of the layerwise error routing topology on the circle dataset as error signal delay and sample duration are swept. The network topologies increase in the number of hidden layers, where each added hidden layer has 24 neurons. From left to right, the networks have one hidden layer of 24 ReLU neurons, two hidden layers of 24 ReLU neurons, and three hidden layers of 24 ReLU neurons. Layerwise error routing imposes stricter requirements on delay due to longer error propagation paths.}
    \label{fig:mlp_heatmaps_delay_hold_circles}
\end{figure}

The layerwise error routing topology propagates error from the output layer step by step to previous layers via $V_l\boldsymbol{\epsilon}_l$. This corresponds to a KP or weight mirroring scheme where the backward weights $V_l$ are learned and converge to transposes of the forward weights. Figure~\ref{fig:mlp_heatmaps_delay_hold_circles} examines the accuracy of the layerwise error routing topology on a synthetic two-dimensional circles dataset. This task is more nonlinear than the $7\times7$ downsampled MNIST task, and requires deeper ReLU networks to represent an accurate decision boundary. 

Unlike the direct error routing topology, the layerwise error routing topology suffers more from finite propagation times: not only does the input signal have to propagate to the output layer, but the error signal must also propagate backward from the output layer back to the input layer. The heatmaps show that these deeper networks require longer sample times and are more sensitive to delay in the error signal. Finite propagation speed means that deeper networks accumulate lag, degrading the learning signal especially in early layers. Longer sample durations mitigate this effect by increasing the window where inputs and their true errors coincide. In the language of Section~\ref{sec:timing-robustness}, this increases $T$ relative to a fixed $\Delta$, resulting in a greater correct update $\mathbb E[\Delta W]$. In the lower triangular regions of the heatmap (large absolute delays), overlap is insufficient and learning fails entirely.

\subsection{Biological Timescales}\label{sec:biological-timescales}
\begin{table}[h!]
\caption{Biophysical timescales in continuous-time network models. Time constants correspond to distinct molecular/cellular mechanisms: rapid receptor/channel kinetics ($\tprop{}$), seconds-scale intracellular signaling and receptor trafficking ($\tplas{}$), and slow homeostatic regulation ($\tdec{}$).}
\centering
\begin{tabularx}{\textwidth}{
    >{\RaggedRight\arraybackslash}p{0.04\textwidth}
    X
    X
    X
}
\toprule
\textbf{} &
\textbf{Corresponding Biophysical Process} &
\textbf{Biological Plausible Range} &
\textbf{Successful Range (Fig~\ref{fig:mlp-heatmaps-tau-pot-inf})} \\
\midrule
$\tprop{}$ 
& Fast synaptic transmission determined by receptor/channel kinetics and membrane RC filtering such as AMPA/GABA$_A$ receptor conductance decay after vesicular glutamate/GABA release \citep{destexhe1998kinetic,o1998activity}. 
& $2$–$30$ms in cortex~\citep{kapur1997gabaa}; sub-ms in auditory brainstem synapses~\citep{geiger1997submillisecond}.
& Swept over 5–30ms, best learning with $\tprop{}<20$ms. \\ 
\addlinespace
$\tplas{}$  
& Coincidence-gated plasticity via second-messenger cascades (dopamine D1/D2$\!\to$cAMP/PKA, Ca$^{2+}\!\to$CaMKII)~\citep{nicoll2023synaptic}, regulating AMPAR phosphorylation and trafficking. Defines the biochemical “induction gate’’ during which pre/post and modulatory signals interact \citep{yagishita2014critical,gerstner2018eligibility}. 
& $\sim$0.3–10s depending on circuit (striatal vs. cortical/hippocampal). Some studies find an effective eligibility trace for dopamine-driven LTP lasts 1–2 minutes, while weaker effects can persist up to 10 minutes~\citep{brzosko2015retroactive}.
& 1.45s to 10s, with best learning in $\tplas{} \gtrsim 2$s. \\
\addlinespace
$\tdec{}$  
& Slow synaptic weakening via protein turnover, phosphatase activity, and homeostatic scaling \citep{turrigiano1998activity, ehlers2003activity} (e.g., AMPAR endocytosis~\citep{ehlers2000reinsertion}, transcriptional regulation).
& Minutes to tens of minutes.
& $\sim 20$ minutes. We find that as long as $\tdec{}$ is sufficiently larger than $\tprop{}, \tplas{}$, it plays little role in the learning dynamics. \\
\bottomrule
\end{tabularx}
\label{tab:timescales}
\end{table}

A central motivation for these algorithms is their relevance to biological learning. Real neural circuits operate on multiple nested timescales, from millisecond synaptic conductances to second-scale plasticity windows and slower homeostatic processes. Yet most algorithmic work has abstracted away these temporal constraints. Here, we parameterize our continuous-time networks with biologically motivated constants and show that they learn effectively on timescales observed in the brain. Of the three time constants relevant in our model, $\tplas{}$ is the least constrained biologically. Our results therefore provide a new theoretical prediction that narrows its plausible functional range. 

A central result of our study is that robust learning requires plasticity windows that outlast the stimulus duration by at least an order of magnitude, placing $\tplas{}$ firmly in the few-second range. This prediction is biologically plausible and experimentally testable, and it narrows the functional range of $\tplas{}$, which has so far been the least constrained timescale in biology. Our simulations demonstrate that the hierarchy $\tprop{}\ll\tplas{}\ll\tdec{}$, characteristic of cortical tissue, is sufficient to support effective learning.

In our model, $\tprop{}$ corresponds to the dominant synaptic conductance time constant, $\tplas{}$ to the biochemical induction gate during which coincident presynaptic drive and modulatory/error input can trigger plasticity, and $\tdec{}$ to slow synaptic weakening. Once $\tdec{}\gg\tprop{},\tplas{}$, it primarily sets a slow baseline for weight decay rather than shaping learning dynamics. Table~\ref{tab:timescales} summarizes these biophysical processes.

\begin{figure}[h!]
    \centering
    \includegraphics[width=1\linewidth]{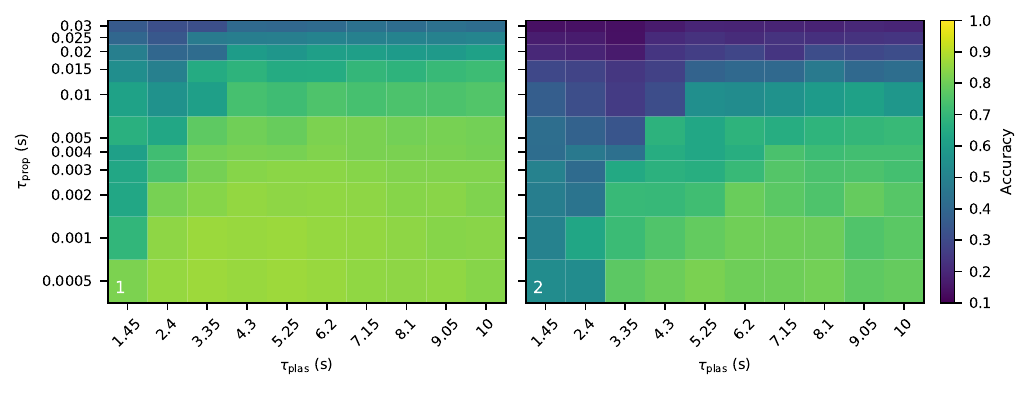}
    \caption{Evaluation of a layerwise error routing network on the $7\times7$ MNIST dataset. 
    \textbf{Left:} 1 hidden layer (49 neurons). \textbf{Right:} 2 hidden layers (49 and 32 neurons). 
    Learning is unstable when the plasticity timescale $\tplas{}$ is comparable to the 
    presentation window ($T=50$\,ms), but becomes robust only once $\tplas{}$ exceeds $\sim2$ s. 
    This corresponds to $\tplas{}/T \approx 40$, highlighting the requirement that plasticity 
    persist far longer than the input presentation time.
    }
    \label{fig:mlp-heatmaps-tau-pot-inf}
\end{figure}

Figure~\ref{fig:mlp-heatmaps-tau-pot-inf} evaluates the trained accuracy of two layerwise error routing networks on these biologically grounded timescales. Learning is unstable when $\tplas{}$ is comparable to the input window ($T=50$ ms), but stabilizes once $\tplas{}\gtrsim2$ s, corresponding to $\tplas{}/T \approx 40$. This finding directly links biological timescales to algorithmic viability, identifying seconds-scale eligibility traces as a necessary condition for feedback-driven learning in cortical-like circuits.

\section{Discussion}
We asked whether feedback-alignment learning can operate in continuous time, without artificial phase separation and at realistic neural timescales. To answer this question, we formulated a coupled ODE model of feedback alignment and characterized its behavior as a function of error delays and plasticity time constants. By embedding neural activity and synaptic plasticity within a single set of differential equations, we show that learning can emerge from continuous-time dynamics without any explicit synchronization.

Our model demonstrates that the central requirement is not weight symmetry, but temporal correlation, which has been an belief in reinforcement learning and neuroscience \citep{gerstner2018eligibility}. Synaptic updates are proportional to the filtered cross-correlation between presynaptic activity and locally projected error signals. When these signals overlap in time, learning approximates gradient descent; when overlap vanishes, updates become biased and performance collapses. In this view, weight transport is not the primary obstacle to biological plausibility. The deeper constraint is maintaining input–error coincidence within the plasticity window of each synapse. 

This perspective clarifies the functional role of timescale separation. The hierarchy ${\tprop{}\ll\tplas{}\ll\tdec{}}$ ensures that neural states track inputs rapidly, plasticity integrates over stimulus windows, and long-term decay does not interfere with short-term credit assignment. Timescale separation acts as a structural condition for stable dynamical learning, independent of the specific feedback topology (FA, DFA, or KP). 

Therefore, within the limits of a rate-based dynamical model, our results suggest that feedback-alignment--style algorithms can be compatible with biological timescales. These results also suggest that it may be difficult to scale such algorithms to deeper networks in continuous time, though this may not be a major problem because it is yet unclear how ``deep'' biological brains actually are, and there certainly exist parts of the human brain that can be modeled by shallow networks~\citep{lillicrap2020backpropagation, akrout2019deep}.

These results could extend beyond biological modeling. Any analog physical substrate must contend with finite propagation delays and heterogeneous device dynamics. Our analysis suggests that such systems need not enforce exact symmetry between forward and backward pathways. Instead, they must preserve sufficient temporal overlap between signals driving synaptic change. Continuous-time learning is therefore feasible in physical systems so long as this overlap condition is met. More broadly, this work connects error-propagation algorithms to the theory of dynamical systems. Learning can be understood as the accumulation of temporally filtered correlations under coupled state–parameter evolution. In this formulation, gradient alignment arises as a consequence of sustained coincidence rather than architectural symmetry. This suggests a unified lens through which to interpret cortical learning, feedback-alignment algorithms, and hardware implementations as instances of the same dynamical principle.

\bibliography{refs}

\appendix

\section{Dynamics of a Single Neuron}\label{sec:appendix-neuron-dynamics}
The single-neuron version of the dynamics presented in~\eqref{eq:layer-dynamics}:
\begin{align*}
    \dot z = \frac{-\,z + \sigma(\mathbf w^\top \mathbf x)}{\tprop}, \quad
    \dot{\mathbf w} = -\frac{\mathbf w}{\tdec{W}} + \frac{(\mathbf v^\top \boldsymbol\epsilon)\,\mathbf x}{\tplas{W}}, \quad
    \dot{\mathbf v} = -\frac{\mathbf v}{\tdec{V}} + \frac{(\mathbf w^\top \mathbf x)\,\boldsymbol\epsilon}{\tplas{V}}.
    \label{eq:neuron-dynamics}
\end{align*}

\section{Discrete-time limits}\label{appendix:discrete-time-limits}
The following discrete-time limits corresponding to SGD, FA, DFA, and KP follow directly from the updates analyzed by \citet{ziyin2025heterosynaptic}. In discrete-time FA/DFA, the local modulatory (learning) drive $V_l\,\boldsymbol\epsilon_l(t)\in \mathbb R^{d_l}$ is often written with an additional factor $\sigma_l'(\mathbf a_l)$ where $\mathbf a_l=W_l\mathbf z_{l-1}$. For the ReLU activation used in our experiments, $\sigma_l'$ is just a binary activity mask, so we treat $V_l\,\boldsymbol\epsilon_l$ as the effective gated drive and do not write $\sigma_l'$ separately. Given the error $\mathbf e_L$ as the gradient of the loss function with respect to the output layer:

\textbf{SGD.} Let $\mathbf a_l = W_l\mathbf z_{l-1}$ and define the backpropagated pre-activation error $\boldsymbol\delta_l := \partial\mathcal L/\partial \mathbf a_l$. Set the output-layer error source to $\boldsymbol\epsilon_L=\boldsymbol\delta_L$, and for hidden layers set $\boldsymbol\epsilon_l=\boldsymbol\delta_{l+1}$ with feedback weights $V_l=W_{l+1}^\top$ (for $l<L$), so that $V_l\boldsymbol\epsilon_l=\boldsymbol\delta_l$ (after the implicit ReLU gate). Then $\Delta W_l\propto \mathbb E\!\left[\boldsymbol\delta_l \,\mathbf z_{l-1}^\top\right]$, which is exactly the gradient of the loss.

\textbf{Feedback Alignment (FA).} Set $V_l$ to fixed random matrices (of shape $d_l\times d_{l+1}$) that project the error from layer $l+1$ to layer $l$. Let $\boldsymbol\epsilon_L=\boldsymbol\delta_L$ and $\boldsymbol\epsilon_l=\boldsymbol\delta_{l+1}$ for $l<L$, so that the local learning drive is $V_l\boldsymbol\epsilon_l$. Then $\Delta W_l\propto \mathbb E\!\left[\big(V_l\boldsymbol\epsilon_l\big)\,\mathbf z_{l-1}^\top\right]$, which is the standard FA update. Gradient alignment arises empirically~\citep{lillicrap2016random}.

\textbf{Direct Feedback Alignment (DFA).} Set $V_l$ to fixed random matrices (of shape $d_l\times d_L$). Broadcast the output-layer error to all hidden layers by setting $\boldsymbol\epsilon_l=\boldsymbol\delta_L$ for all $l$. Then the local learning drive is $V_l\boldsymbol\delta_L$ and
$\Delta W_l\propto \mathbb E\!\left[\big(V_l\boldsymbol\delta_L\big)\,\mathbf z_{l-1}^\top\right]$, which is the DFA update: each layer learns from a direct projection of the global error~\citep{nokland2016direct}.

\textbf{Kolen--Pollack (KP) / Weight-Mirror Methods.} Allow $V_l$ to evolve under the full plasticity rules in~\eqref{eq:layer-dynamics}. In the layerwise error-routing topology we take $\boldsymbol\epsilon_L=\boldsymbol\delta_L$ and, for $l<L$, $\boldsymbol\epsilon_l=\boldsymbol\delta_{l+1}$ so that $V_l$ is driven by correlations between the forward drive $W_l\mathbf z_{l-1}$ and the next-layer error source. In expectation, this update combined with decay drives $V_l$ toward a transpose-like mapping that approximates $W_{l+1}^\top$, recovering a backprop-like learning drive $V_l\boldsymbol\epsilon_l$ without enforcing weight transport~\citep{akrout2019deep, kolen1994backpropagation}.

\section{Datasets \& Methodology}
We use two datasets in our evaluation of these models. The $7\times7$ downsampled MNIST dataset is the standard MNIST~\citep{lecun1998mnist} dataset that has been downsampled with $4\times4$ average pooling. The circles dataset is the \texttt{make\_circles} dataset from scikit-learn~\citep{pedregosa2011scikit}. It is well established that simple linear classifiers achieve surprisingly high accuracy on MNIST, with only modest improvements from deeper architectures~\citep{lecun2002gradient}. We find a linear softmax regressor on $7\times7$ downsampled MNIST achieves over 89\% test accuracy. In contrast, a logistic regressor on the 2-circle concentric-rings dataset achieves only 75\% test accuracy, which corresponds to the class imbalance itself (75\% outer ring, 25\% inner ring).

$W$ are initialized according Xavier normalization~\citep{glorot2010understanding}. $V$ are initialized to a fixed constant 0.1. Classification decisions are read out from output neurons at the very end of each input sample's presentation. Evaluation is done on frozen $W$ and $V$ dynamics. Each heatmap data point is the average of 3-5 runs, depending on the experiment. 

We integrate all continuous-time dynamics using Diffrax's~\citep{kidger2021on} Tsit5 solver~\citep{tsitouras2011runge} (a fifth-order explicit Runge--Kutta method with an embedded fourth-order error estimate) equipped with a PID adaptive step-size controller (\texttt{rtol} $= 2\times 10^{-3}$, \texttt{atol} $= 10^{-5}$). This setup allows the solver to take large steps during slowly varying segments of the dynamics while automatically refining steps around rapid transients induced by input switches and error onsets. Although the timescales of our systems span several orders of magnitude, we empirically find that they are not so stiff as to require an implicit method; Tsit5 remains stable and efficient under these tolerances. Compared to a fixed-step forward Euler integrator tuned to resolve the fastest timescale, this adaptive scheme reduces wall-clock time by several orders of magnitude while producing indistinguishable learning curves and dynamics.

\section{Comparison to SGD}
To ensure a fair comparison, we train baseline multi-layer perceptrons (MLPs) with the same hidden unit counts as our continuous-time networks with layerwise error propagation. Both models are exposed to the same training data samples in the same order. For the baselines, we use the Adam optimizer~\citep{kingma2014adam} with an initial learning rate of $0.001$, rather than vanilla SGD, to avoid penalizing the baseline with suboptimal hyperparameters. Training is performed with a batch size of 1, matching the presentation schedule used by the continuous-time networks. All networks use ReLU activations~\citep{glorot2011deep} on hidden layers. 

Table~\ref{tab:mlp-accuracy} shows accuracies on the $7\times7$ MNIST dataset. In shallow networks, our layerwise error propagation network matches the baseline on both training and test accuracy, but in deeper networks lags slightly. On simple and fairly linear tasks, this is expected: signal lag accumulates with depth (Section~\ref{sec:timing-robustness}), and the dataset itself yields diminishing returns from additional layers.
\begin{table}[h]
\centering
\begin{tabular}{lcc|cc}
\toprule
& \multicolumn{2}{c}{\textbf{Train Accuracy}} & \multicolumn{2}{c}{\textbf{Test Accuracy}} \\
\cmidrule(lr){2-3} \cmidrule(lr){4-5}
\textbf{Model} & Discrete-time & Continuous-time & Discrete-time & Continuous-time \\
\midrule
1-layer & $0.8750 \pm 0.0054$ & $\mathbf{0.9318 \pm 0.0028}$ & $0.8634 \pm 0.0056$ & $\mathbf{0.9209 \pm 0.0011}$ \\
2-layer & $0.8758 \pm 0.0220$ & $\mathbf{0.9318 \pm 0.0032}$ & $0.8673 \pm 0.0259$ & $\mathbf{0.8919 \pm 0.0050}$ \\
\bottomrule
\end{tabular}
\caption{Accuracy (mean $\pm$ std) on the $7\times7$ MNIST dataset across 3 runs. We compare discrete-time MLP baselines (SGD with Adam, lr$=0.001$) to our continuous-time layerwise error propagation network across architectures with 1–2 hidden layers.}
\label{tab:mlp-accuracy}
\end{table}

On the more nonlinear task of distinguishing concentric circles, additional depth provides a benefit (Table~\ref{tab:accuracy-circles}). Our continuous-time layerwise error propagation networks achieve comparable or higher test accuracies than the discrete-time baselines.
\begin{table}[h]
\centering
\begin{tabular}{lcc|cc}
\toprule
& \multicolumn{2}{c}{\textbf{Train Accuracy}} & \multicolumn{2}{c}{\textbf{Test Accuracy}} \\
\cmidrule(lr){2-3} \cmidrule(lr){4-5}
\textbf{Model} & Discrete-time & Continuous-time & Discrete-time & Continuous-time \\
\midrule
1-layer & $0.8318 \pm 0.0276$ & $\mathbf{0.9241 \pm 0.0133}$ & $0.8272 \pm 0.0282$ & $\mathbf{0.9470 \pm 0.0101}$ \\
2-layer & $0.9520 \pm 0.0076$ & $\mathbf{0.9657 \pm 0.0082}$ & $0.9485 \pm 0.0066$ & $\mathbf{0.9683 \pm 0.0109}$ \\
3-layer & $0.9322 \pm 0.0300$ & $\mathbf{0.9494 \pm 0.0588}$ & $0.9312 \pm 0.0218$ & $\mathbf{0.9568 \pm 0.0452}$ \\
\bottomrule
\end{tabular}
\caption{Accuracy (mean $\pm$ std) on the concentric circles dataset across 3 runs. We compare discrete-time MLP baselines (SGD with Adam, lr $=0.001$) to our continuous-time layerwise error propagation network across architectures with 1–3 hidden layers.}
\label{tab:accuracy-circles}
\end{table}

We can also compare the resilience of our models versus standard SGD in the presence of label error. In our continuous time networks, this arises due to delays in the error signal, as discussed in Section~\ref{sec:timing-robustness}. We can model the same error in standard discrete-time neural networks by presenting a standard neural network optimizer a corresponding fraction of mislabeled data. For example, a delay of half the sample time, $\Delta = \frac12T$ in our continuous-time model, corresponds to presenting the correct data pairing $(X_i,Y_i)$ and then the mismatched label $(X_i, Y_{i+1})$, where $i$ is the index into the input and label datasets. 

\begin{figure}
    \centering
    \includegraphics[width=\linewidth]{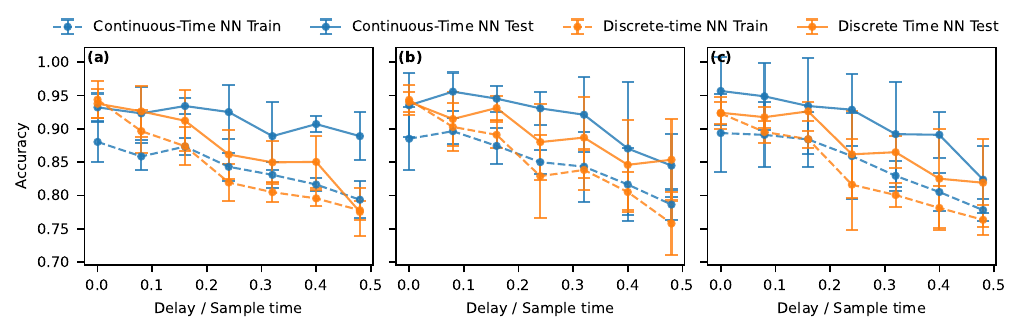}
    \caption{Delay ratio versus accuracy. Panels (a), (b), and (c) correspond to networks with 1, 2, and 3 hidden layers. Our continuous-time neural networks with the layerwise error propagation topology demonstrate comparable, if not slightly better, robustness to mismatch than corresponding discrete-time neural networks.}
    \label{fig:delay-ratio-vs-accuracy}
\end{figure}

Figure~\ref{fig:delay-ratio-vs-accuracy} compares our continuous-time neural network with a standard discrete-time neural network trained with \emph{label dithering}, which models the effect of delayed error signals without duplicating training data. Instead of splitting each sample into sub-intervals, we approximate the delay ratio \(\,r = \tau_{\text{delay}} / T_{\text{sample}}\,\) by a rational fraction \(p/q\), and then assign exactly \(p\) out of every \(q\) samples to use the \emph{previous} label while the rest use the current label. This ensures that the fraction of ``mismatched" updates matches the physical delay ratio in expectation, while keeping each sample presented only once. In this way, the training dynamics reflect the temporal overlap of input and delayed error, 
but the dataset size and number of SGD steps remain unchanged.

\end{document}

%% file: math_commands.tex
\usepackage{amsmath,amsfonts,bm}

\def\eqref#1{equation~\ref{#1}}

\def\1{\bm{1}}

\DeclareMathAlphabet{\mathsfit}{\encodingdefault}{\sfdefault}{m}{sl}
\SetMathAlphabet{\mathsfit}{bold}{\encodingdefault}{\sfdefault}{bx}{n}



%% file: figures/topologies.tex
\tikzset{every picture/.style={line width=0.75pt}} 

\begin{tikzpicture}[x=0.75pt,y=0.75pt,yscale=-1,xscale=1]

\draw   (205.8,114.15) .. controls (205.8,111.33) and (208.08,109.05) .. (210.9,109.05) -- (270.32,109.05) .. controls (273.13,109.05) and (275.41,111.33) .. (275.41,114.15) -- (275.41,129.44) .. controls (275.41,132.26) and (273.13,134.54) .. (270.32,134.54) -- (210.9,134.54) .. controls (208.08,134.54) and (205.8,132.26) .. (205.8,129.44) -- cycle ;
\draw   (205.16,165.76) .. controls (205.16,162.94) and (207.45,160.66) .. (210.26,160.66) -- (270.32,160.66) .. controls (273.13,160.66) and (275.41,162.94) .. (275.41,165.76) -- (275.41,181.05) .. controls (275.41,183.87) and (273.13,186.15) .. (270.32,186.15) -- (210.26,186.15) .. controls (207.45,186.15) and (205.16,183.87) .. (205.16,181.05) -- cycle ;
\draw   (205.16,216.1) .. controls (205.16,213.28) and (207.45,211) .. (210.26,211) -- (270.32,211) .. controls (273.13,211) and (275.41,213.28) .. (275.41,216.1) -- (275.41,231.39) .. controls (275.41,234.21) and (273.13,236.49) .. (270.32,236.49) -- (210.26,236.49) .. controls (207.45,236.49) and (205.16,234.21) .. (205.16,231.39) -- cycle ;
\draw    (240.37,160.66) -- (240.37,136.54) ;
\draw [shift={(240.37,134.54)}, rotate = 90] [color={rgb, 255:red, 0; green, 0; blue, 0 }  ][line width=0.75]    (10.93,-3.29) .. controls (6.95,-1.4) and (3.31,-0.3) .. (0,0) .. controls (3.31,0.3) and (6.95,1.4) .. (10.93,3.29)   ;
\draw    (240.37,211) -- (240.37,187.51) ;
\draw [shift={(240.37,185.51)}, rotate = 90] [color={rgb, 255:red, 0; green, 0; blue, 0 }  ][line width=0.75]    (10.93,-3.29) .. controls (6.95,-1.4) and (3.31,-0.3) .. (0,0) .. controls (3.31,0.3) and (6.95,1.4) .. (10.93,3.29)   ;
\draw   (389.69,232.54) .. controls (389.7,235.35) and (387.42,237.64) .. (384.61,237.65) -- (325.19,237.87) .. controls (322.37,237.88) and (320.08,235.61) .. (320.07,232.79) -- (320.02,217.5) .. controls (320.01,214.69) and (322.28,212.4) .. (325.1,212.39) -- (384.51,212.17) .. controls (387.33,212.16) and (389.62,214.43) .. (389.63,217.25) -- cycle ;
\draw   (390.13,180.92) .. controls (390.14,183.74) and (387.87,186.03) .. (385.06,186.04) -- (325,186.26) .. controls (322.19,186.27) and (319.89,184) .. (319.88,181.18) -- (319.83,165.89) .. controls (319.82,163.07) and (322.09,160.78) .. (324.91,160.77) -- (384.96,160.55) .. controls (387.78,160.54) and (390.07,162.82) .. (390.08,165.63) -- cycle ;
\draw   (389.95,130.59) .. controls (389.96,133.4) and (387.69,135.69) .. (384.87,135.7) -- (324.82,135.92) .. controls (322,135.93) and (319.71,133.66) .. (319.7,130.84) -- (319.64,115.55) .. controls (319.63,112.74) and (321.91,110.45) .. (324.72,110.44) -- (384.78,110.21) .. controls (387.59,110.2) and (389.88,112.48) .. (389.89,115.29) -- cycle ;
\draw    (354.95,186.15) -- (355.04,210.28) ;
\draw [shift={(355.04,212.28)}, rotate = 269.79] [color={rgb, 255:red, 0; green, 0; blue, 0 }  ][line width=0.75]    (10.93,-3.29) .. controls (6.95,-1.4) and (3.31,-0.3) .. (0,0) .. controls (3.31,0.3) and (6.95,1.4) .. (10.93,3.29)   ;
\draw    (355.4,135.18) -- (355.49,159.3) ;
\draw [shift={(355.49,161.3)}, rotate = 269.79] [color={rgb, 255:red, 0; green, 0; blue, 0 }  ][line width=0.75]    (10.93,-3.29) .. controls (6.95,-1.4) and (3.31,-0.3) .. (0,0) .. controls (3.31,0.3) and (6.95,1.4) .. (10.93,3.29)   ;
\draw    (277.41,122.43) -- (317.86,122.43) ;
\draw [shift={(319.86,122.43)}, rotate = 180] [color={rgb, 255:red, 0; green, 0; blue, 0 }  ][line width=0.75]    (10.93,-3.29) .. controls (6.95,-1.4) and (3.31,-0.3) .. (0,0) .. controls (3.31,0.3) and (6.95,1.4) .. (10.93,3.29)   ;
\draw [shift={(275.41,122.43)}, rotate = 0] [color={rgb, 255:red, 0; green, 0; blue, 0 }  ][line width=0.75]    (10.93,-3.29) .. controls (6.95,-1.4) and (3.31,-0.3) .. (0,0) .. controls (3.31,0.3) and (6.95,1.4) .. (10.93,3.29)   ;
\draw    (277.41,173.41) -- (317.86,173.41) ;
\draw [shift={(319.86,173.41)}, rotate = 180] [color={rgb, 255:red, 0; green, 0; blue, 0 }  ][line width=0.75]    (10.93,-3.29) .. controls (6.95,-1.4) and (3.31,-0.3) .. (0,0) .. controls (3.31,0.3) and (6.95,1.4) .. (10.93,3.29)   ;
\draw [shift={(275.41,173.41)}, rotate = 0] [color={rgb, 255:red, 0; green, 0; blue, 0 }  ][line width=0.75]    (10.93,-3.29) .. controls (6.95,-1.4) and (3.31,-0.3) .. (0,0) .. controls (3.31,0.3) and (6.95,1.4) .. (10.93,3.29)   ;
\draw    (277.41,224.38) -- (317.86,224.38) ;
\draw [shift={(319.86,224.38)}, rotate = 180] [color={rgb, 255:red, 0; green, 0; blue, 0 }  ][line width=0.75]    (10.93,-3.29) .. controls (6.95,-1.4) and (3.31,-0.3) .. (0,0) .. controls (3.31,0.3) and (6.95,1.4) .. (10.93,3.29)   ;
\draw [shift={(275.41,224.38)}, rotate = 0] [color={rgb, 255:red, 0; green, 0; blue, 0 }  ][line width=0.75]    (10.93,-3.29) .. controls (6.95,-1.4) and (3.31,-0.3) .. (0,0) .. controls (3.31,0.3) and (6.95,1.4) .. (10.93,3.29)   ;
\draw    (240.37,109.05) -- (240.37,84.93) ;
\draw [shift={(240.37,82.93)}, rotate = 90] [color={rgb, 255:red, 0; green, 0; blue, 0 }  ][line width=0.75]    (10.93,-3.29) .. controls (6.95,-1.4) and (3.31,-0.3) .. (0,0) .. controls (3.31,0.3) and (6.95,1.4) .. (10.93,3.29)   ;
\draw    (240.37,262.61) -- (240.37,238.49) ;
\draw [shift={(240.37,236.49)}, rotate = 90] [color={rgb, 255:red, 0; green, 0; blue, 0 }  ][line width=0.75]    (10.93,-3.29) .. controls (6.95,-1.4) and (3.31,-0.3) .. (0,0) .. controls (3.31,0.3) and (6.95,1.4) .. (10.93,3.29)   ;
\draw    (354.76,84.2) -- (354.85,108.32) ;
\draw [shift={(354.86,110.32)}, rotate = 269.79] [color={rgb, 255:red, 0; green, 0; blue, 0 }  ][line width=0.75]    (10.93,-3.29) .. controls (6.95,-1.4) and (3.31,-0.3) .. (0,0) .. controls (3.31,0.3) and (6.95,1.4) .. (10.93,3.29)   ;
\draw   (430.55,115.78) .. controls (430.55,112.97) and (432.84,110.69) .. (435.65,110.69) -- (495.07,110.69) .. controls (497.88,110.69) and (500.17,112.97) .. (500.17,115.78) -- (500.17,131.08) .. controls (500.17,133.89) and (497.88,136.18) .. (495.07,136.18) -- (435.65,136.18) .. controls (432.84,136.18) and (430.55,133.89) .. (430.55,131.08) -- cycle ;
\draw   (429.92,167.4) .. controls (429.92,164.58) and (432.2,162.3) .. (435.01,162.3) -- (495.07,162.3) .. controls (497.88,162.3) and (500.17,164.58) .. (500.17,167.4) -- (500.17,182.69) .. controls (500.17,185.51) and (497.88,187.79) .. (495.07,187.79) -- (435.01,187.79) .. controls (432.2,187.79) and (429.92,185.51) .. (429.92,182.69) -- cycle ;
\draw   (429.92,217.74) .. controls (429.92,214.92) and (432.2,212.64) .. (435.01,212.64) -- (495.07,212.64) .. controls (497.88,212.64) and (500.17,214.92) .. (500.17,217.74) -- (500.17,233.03) .. controls (500.17,235.84) and (497.88,238.13) .. (495.07,238.13) -- (435.01,238.13) .. controls (432.2,238.13) and (429.92,235.84) .. (429.92,233.03) -- cycle ;
\draw    (465.12,162.3) -- (465.12,138.18) ;
\draw [shift={(465.12,136.18)}, rotate = 90] [color={rgb, 255:red, 0; green, 0; blue, 0 }  ][line width=0.75]    (10.93,-3.29) .. controls (6.95,-1.4) and (3.31,-0.3) .. (0,0) .. controls (3.31,0.3) and (6.95,1.4) .. (10.93,3.29)   ;
\draw    (465.12,212.64) -- (465.12,189.15) ;
\draw [shift={(465.12,187.15)}, rotate = 90] [color={rgb, 255:red, 0; green, 0; blue, 0 }  ][line width=0.75]    (10.93,-3.29) .. controls (6.95,-1.4) and (3.31,-0.3) .. (0,0) .. controls (3.31,0.3) and (6.95,1.4) .. (10.93,3.29)   ;
\draw    (465.12,110.69) -- (465.12,86.56) ;
\draw [shift={(465.12,84.56)}, rotate = 90] [color={rgb, 255:red, 0; green, 0; blue, 0 }  ][line width=0.75]    (10.93,-3.29) .. controls (6.95,-1.4) and (3.31,-0.3) .. (0,0) .. controls (3.31,0.3) and (6.95,1.4) .. (10.93,3.29)   ;
\draw    (465.12,264.25) -- (465.12,240.13) ;
\draw [shift={(465.12,238.13)}, rotate = 90] [color={rgb, 255:red, 0; green, 0; blue, 0 }  ][line width=0.75]    (10.93,-3.29) .. controls (6.95,-1.4) and (3.31,-0.3) .. (0,0) .. controls (3.31,0.3) and (6.95,1.4) .. (10.93,3.29)   ;
\draw    (545.74,83.93) -- (545.84,212) ;
\draw    (502.39,123.09) .. controls (518.68,122.76) and (545.43,118.87) .. (545.84,110.05) ;
\draw [shift={(500.17,123.11)}, rotate = 0] [color={rgb, 255:red, 0; green, 0; blue, 0 }  ][line width=0.75]    (10.93,-3.29) .. controls (6.95,-1.4) and (3.31,-0.3) .. (0,0) .. controls (3.31,0.3) and (6.95,1.4) .. (10.93,3.29)   ;
\draw    (501.76,175.33) .. controls (518.06,174.87) and (545,169.53) .. (545.41,160.71) ;
\draw [shift={(499.53,175.36)}, rotate = 0] [color={rgb, 255:red, 0; green, 0; blue, 0 }  ][line width=0.75]    (10.93,-3.29) .. controls (6.95,-1.4) and (3.31,-0.3) .. (0,0) .. controls (3.31,0.3) and (6.95,1.4) .. (10.93,3.29)   ;
\draw    (502.39,225.04) .. controls (518.68,224.71) and (545.43,220.82) .. (545.84,212) ;
\draw [shift={(500.17,225.06)}, rotate = 0] [color={rgb, 255:red, 0; green, 0; blue, 0 }  ][line width=0.75]    (10.93,-3.29) .. controls (6.95,-1.4) and (3.31,-0.3) .. (0,0) .. controls (3.31,0.3) and (6.95,1.4) .. (10.93,3.29)   ;
\draw   (54.25,177.49) .. controls (54.25,165.58) and (63.91,155.91) .. (75.83,155.91) .. controls (87.75,155.91) and (97.41,165.58) .. (97.41,177.49) .. controls (97.41,189.41) and (87.75,199.08) .. (75.83,199.08) .. controls (63.91,199.08) and (54.25,189.41) .. (54.25,177.49) -- cycle ;
\draw    (75.83,238.57) -- (75.83,201.08) ;
\draw [shift={(75.83,199.08)}, rotate = 90] [color={rgb, 255:red, 0; green, 0; blue, 0 }  ][line width=0.75]    (10.93,-3.29) .. controls (6.95,-1.4) and (3.31,-0.3) .. (0,0) .. controls (3.31,0.3) and (6.95,1.4) .. (10.93,3.29)   ;
\draw    (75.83,155.91) -- (75.83,126.62) ;
\draw [shift={(75.83,124.62)}, rotate = 90] [color={rgb, 255:red, 0; green, 0; blue, 0 }  ][line width=0.75]    (10.93,-3.29) .. controls (6.95,-1.4) and (3.31,-0.3) .. (0,0) .. controls (3.31,0.3) and (6.95,1.4) .. (10.93,3.29)   ;
\draw  [dash pattern={on 1.5pt off 0.75pt}]  (149.21,156.56) .. controls (128.91,171.79) and (125.19,174.58) .. (99.04,177.33) ;
\draw [shift={(97.41,177.49)}, rotate = 354.2] [color={rgb, 255:red, 0; green, 0; blue, 0 }  ][line width=0.75]    (10.93,-3.29) .. controls (6.95,-1.4) and (3.31,-0.3) .. (0,0) .. controls (3.31,0.3) and (6.95,1.4) .. (10.93,3.29)   ;
\draw  [color={rgb, 255:red, 155; green, 155; blue, 155 }  ,draw opacity=1 ] (10.25,182.12) .. controls (10.25,135.61) and (47.95,97.91) .. (94.46,97.91) .. controls (140.97,97.91) and (178.67,135.61) .. (178.67,182.12) .. controls (178.67,228.63) and (140.97,266.33) .. (94.46,266.33) .. controls (47.95,266.33) and (10.25,228.63) .. (10.25,182.12) -- cycle ;
\draw  [color={rgb, 255:red, 155; green, 155; blue, 155 }  ,draw opacity=1 ] (208,173.38) .. controls (208,167.92) and (212.42,163.5) .. (217.88,163.5) .. controls (223.33,163.5) and (227.75,167.92) .. (227.75,173.38) .. controls (227.75,178.83) and (223.33,183.25) .. (217.88,183.25) .. controls (212.42,183.25) and (208,178.83) .. (208,173.38) -- cycle ;
\draw [color={rgb, 255:red, 155; green, 155; blue, 155 }  ,draw opacity=1 ]   (137.25,109.5) -- (223.25,164.5) ;
\draw [color={rgb, 255:red, 155; green, 155; blue, 155 }  ,draw opacity=1 ]   (145.25,249.5) -- (225.25,179.5) ;
\draw [color={rgb, 255:red, 155; green, 155; blue, 155 }  ,draw opacity=1 ]   (411,15) -- (411,276.5) ;

\draw (355.44,72.83) node   [align=left] {error};
\draw (240.74,72.83) node   [align=left] {output};
\draw (240.11,269.62) node   [align=left] {input};
\draw (465.49,270.62) node   [align=left] {input};
\draw (466.13,73.19) node   [align=left] {output};
\draw (545.78,73.19) node   [align=left] {error};
\draw (77.98,218.44) node [anchor=north west][inner sep=0.75pt]   [align=left] {$\displaystyle \mathbf{w}$};
\draw (71.41,110.12) node [anchor=north west][inner sep=0.75pt]   [align=left] {$\displaystyle z$};
\draw (125.73,175.73) node [anchor=north west][inner sep=0.75pt]   [align=left] {$\displaystyle \mathbf{v}$};
\draw (70.55,244.34) node [anchor=north west][inner sep=0.75pt]   [align=left] {$\displaystyle \mathbf{x}$};
\draw (150.84,150.38) node [anchor=north west][inner sep=0.75pt]   [align=left] {$\displaystyle \mathbf{e}$};
\draw (200,25) node [anchor=north west][inner sep=0.75pt]   [align=left] {\textbf{Layerwise Error Propagation}};
\draw (422,26) node [anchor=north west][inner sep=0.75pt]   [align=left] {\textbf{Direct Error Propagation}};

\end{tikzpicture}